\documentclass{article} 
\usepackage{iclr2021_conference,times}

\usepackage{hyperref}
\usepackage{url}
\usepackage{amsmath}

\usepackage{graphicx,wrapfig,lipsum,multirow,caption,subcaption,siunitx,booktabs}

\usepackage{tabularx}
\newcolumntype{P}[1]{>{\centering\arraybackslash}p{#1}}
\newcolumntype{R}{>{\raggedleft\arraybackslash}X}
\newcolumntype{C}{>{\centering\arraybackslash}X}

\title{Finite Volume Neural Network: Modeling Subsurface Contaminant Transport}

\author{Timothy Praditia \thanks{Corresponding author: \texttt{timothy.praditia@iws.uni-stuttgart.de}} \\
University of Stuttgart \\
\And
Matthias Karlbauer \\
University of T{\"u}bingen \\
\And
Sebastian Otte \\
University of T{\"u}bingen \\
\AND
Sergey Oladyshkin\\
University of Stuttgart \\
\And
Martin V. Butz \\
University of T{\"u}bingen \\
\And
Wolfgang Nowak \\
University of Stuttgart \\
}

\iclrfinalcopy 
\begin{document}

\maketitle

\begin{abstract}

Data-driven modeling of spatiotemporal physical processes with general deep learning methods is a highly challenging task. It is further exacerbated by the limited availability of data, leading to poor generalizations in standard neural network models. To tackle this issue, we introduce a new approach called the Finite Volume Neural Network (FINN). The FINN method adopts the numerical structure of the well-known Finite Volume Method for handling partial differential equations, so that each quantity of interest follows its own adaptable conservation law, while it concurrently accommodates learnable parameters. As a result, FINN enables better handling of fluxes between control volumes and therefore proper treatment of different types of numerical boundary conditions. We demonstrate the effectiveness of our approach with a subsurface contaminant transport problem, which is governed by a non-linear diffusion-sorption process. FINN does not only generalize better to differing boundary conditions compared to other methods, it is also capable to explicitly extract and learn the constitutive relationships (expressed by the retardation factor). More importantly, FINN shows excellent generalization ability when applied to both synthetic datasets and real, sparse experimental data, thus underlining its relevance as a data-driven modeling tool.

\end{abstract}

\section{Introduction}

Training neural networks augmented with additional physical information has been shown to improve their generalization capabilities, particularly when predicting physical processes. In the Physics Informed Neural Network (PINN) framework \citep{Karpatne2017,Karpatne2018,Tartakovsky2018,Raissi2019,Wang2020}, the neural network prediction $u(x,t)$ is defined to be an explicit function of space $x$ and time $t$. Furthermore, calculations of respective derivatives, such as $\smash{\frac{\partial u}{\partial x}}$ and $\smash{\frac{\partial u}{\partial t}}$, are required for formulating the loss function. However, when the available training data is concentrated on a single location $x$ or time $t$, the approximation of the derivatives $\smash{\frac{\partial u}{\partial x}}$ and $\smash{\frac{\partial u}{\partial t}}$ in current techniques deteriorates due to (a) insufficient information provided by the data and (b) the lack of structural explainability of the framework itself. To address these issues from a structural point of view, several works have been conducted in the literature recently. One architecture, namely the Distributed Spatiotemporal Artificial Neural Network \citep[DISTANA,][]{Karlbauer2019}, uses translational invariant Prediction Kernels (PKs) and Transition Kernels (TKs) to process the temporal and spatial correlation of the data, respectively. Another method, called the Universal Differential Equation method \citep[UDE,][]{Rackauckas2020}, combines Convolutional Neural Networks \citep[CNNs,][]{lecun1999object} with Neural Ordinary Differential Equations \citep[NODEs,][]{chen2018neural}, for learning spatiotemporal data. Despite promising results shown by these methods, they still suffer from unreliable flux handling (i.e. the physical fluxes are not guaranteed to be conservative). Consequently, the methods mentioned above lack means to properly treat different types of boundary conditions.

To handle fluxes more robustly and improve generalization within a physical domain, we propose a Finite Volume Neural Network (FINN). The FINN method is a hybrid model, which capitalizes on the structural knowledge of the well-known Finite Volume Method \citep[FVM,][]{Moukalled2016}, and the flexibility as well as the learning abilities of Artificial Neural Networks (ANNs), more specifically NODEs. As a consequence, the FINN structure can properly treat different types of boundary conditions and ensures conservation of the quantities of interest. 
Moreover, we show that FINN is able to reconstruct the full field solution (for all $x$ and $t$) even when trained with only partial information (e.g. at a single point $x$ or $t$). Additionally, the structure of FINN facilitates learning and extracting constitutive relationships and/or reaction terms, and, consequently, shows exceptional generalization capabilities and develops explainable knowledge structures.

\section{Methods}

In this work, we focus on modeling spatiotemporal physical processes, namely processes that scientists try to model with Partial Differential Equations (PDEs), such as diffusion-type problems. They can be generally written mathematically as follows:
\begin{equation}
    \label{eq:diffusion}
    \frac{\partial u}{\partial t} = \frac{\partial}{\partial x} \left(D(u) \frac{\partial u}{\partial x}\right) + q(u),
\end{equation}
where $u$ is the quantity of interest, $t$ is time, $D$ is the diffusion coefficient and $q$ is the source/sink term. Usually, the FVM discretizes \autoref{eq:diffusion} implicitly in space and explicitly in time, leading to a simplified definition:
\begin{equation}
    \label{eq:diffusionFVM}
    \frac{\partial u_i^{(t+1)}}{\partial t} = f\left(u_{i-1}^{(t)},u_{i}^{(t)},u_{i+1}^{(t)},t\right),
\end{equation}
where $u_{i}^{(t)}$ denotes $u$ at control volume $i$ and time step $t$. In other words, the time derivative of $u$ depends on the current value of $u$, the values of $u$ at the neighboring control volumes, and time. For brevity, we drop the time index in later definitions.

In FINN, we introduce Flux Kernels $\mathcal{F}$, which take the input of $u_{i-1}$, $u_{i}$, and $u_{i+1}$ to approximate the divergence part (first term on the right hand side) in \autoref{eq:diffusion} for each control volume $i$:
\begin{equation}
    \label{eq:flux_kernels}
    \mathcal{F}_i = \Phi_{\theta}(u_{i-1}, u_{i}, u_{i+1}) = \sum_s fk_s \approx \oint_S \left(D(u) \frac{\partial u}{\partial x} \cdot \hat{n}\right) \,ds.
\end{equation}


\begin{figure}[b!]
    \centering
    \includegraphics[width=0.7\textwidth]{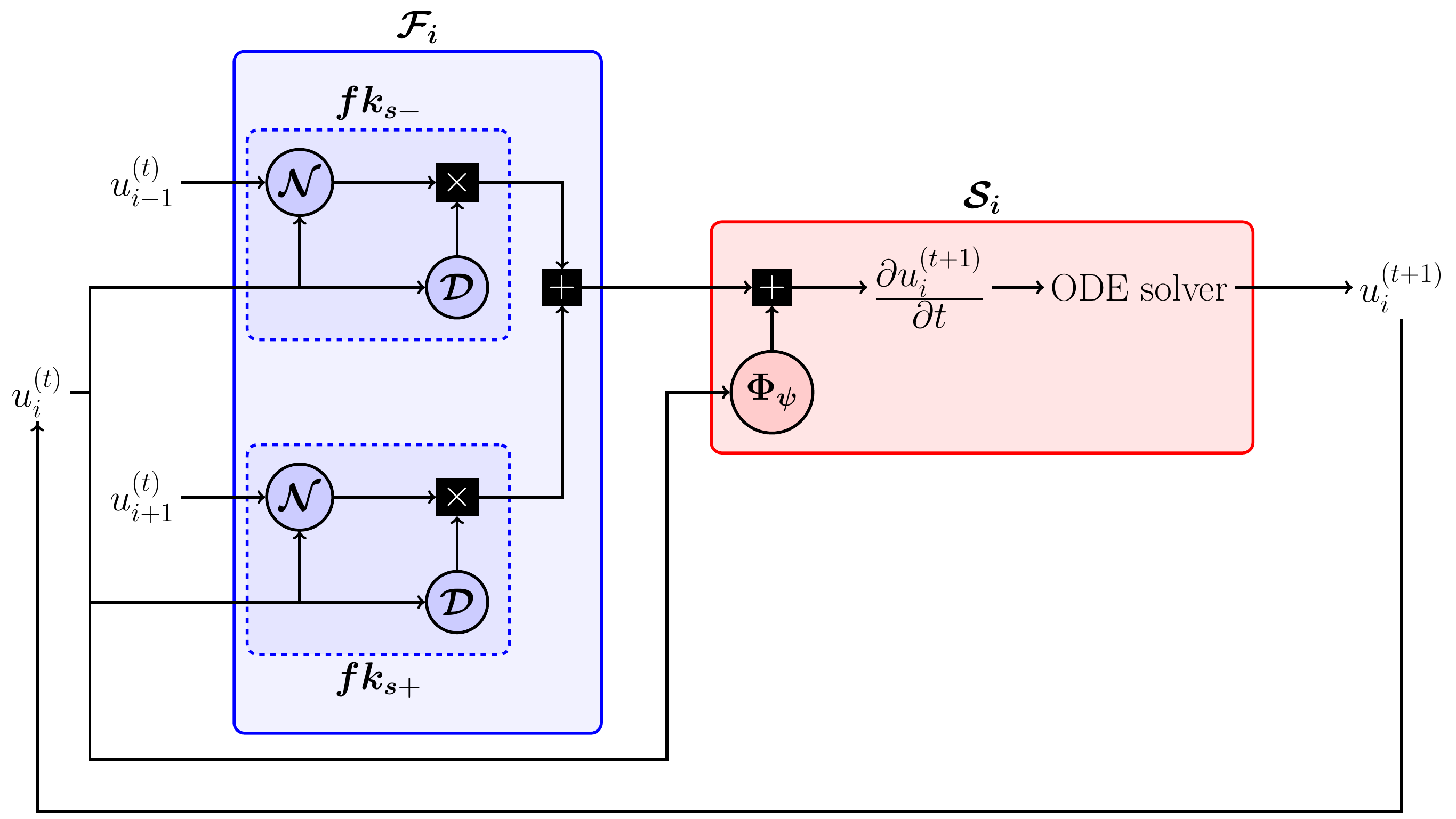}
    \caption{Illustration of the Flux and State Kernels in the FINN.}
    \label{fig:finn}
\end{figure}

The Flux Kernels $\mathcal{F}_i$ consist of subkernels $fk_s$ that calculate fluxes at the boundary surfaces $s$ between control volume $i$ and its neighboring control volumes (see \autoref{fig:finn}), being learned by $\mathcal{N}$ and $\mathcal{D}$, which are subcomponents of the function $\Phi$ with parameters $\theta$. The function $\mathcal{N}$ in each $fk_s$ is equivalent to a linear layer that takes $u_i$ and one of its neighbors (i.e. $u_{i-1}$ or $u_{i+1}$) as inputs and learns the numerical FVM stencil with its weights that should amount to $[-1, 1]$, which corresponds to $[u_i, u_{i-1}]$ and $[u_i, u_{i+1}]$ (i.e. simple difference between neighboring control volumes) in ideal one-dimensional diffusion problems. If the diffusion coefficient $D$ depends on $u$ according to a hidden constitutive relationship, the function $\mathcal{D}$ in each $fk_s$ learns and approximates this function $\mathcal{D}(u) \approx D(u)$. Otherwise, $\mathcal{D}$ will be only a scalar value $\mathcal{D} \equiv D$, which can also be set as a learnable parameter. Next, the output of $\mathcal{N}$ is multiplied with the output of $\mathcal{D}$ to obtain the flux approximation at each boundary. When the fluxes at all boundary surfaces $s$ are integrated in each control volume $i$, the summation of the numerical stencil will lead to the classical one-dimensional numerical Laplacian with $[1, -2, 1]$ corresponding to $[u_{i-1}, u_i, u_{i+1}]$ if \autoref{eq:diffusion} is true.

This structure of the Flux Kernel enables straightforward handling of different types of boundary conditions. With a Dirichlet boundary condition $u = u_b$, we can set either $u_{i-1} = u_b$ or $u_{i+1} = u_b$ at the corresponding domain boundary. With a Neumann boundary condition $\nu$, we can easily set the output of $fk_s$ at the corresponding domain boundary $s$ to be equal to $\nu$. With a Cauchy boundary condition, we can calculate the derivative approximation at the corresponding domain boundary and set it as either $u_{i-1}$ or $u_{i+1}$.


The State Kernels $\mathcal{S}$ then take the output of the Flux Kernels and approximate $\frac{\partial u}{\partial t}$ while also learning the source/sink term $q \approx \Phi_{\psi}(u)$ as a function of $u$ (whenever necessary), which is also learnable by the State Kernels. Formally, each State Kernel can be written as an approximation of the time derivative in \autoref{eq:diffusion} for each control volume $i$:
\begin{equation}
    \label{eq:state_kernels}
    \mathcal{S}_i = \mathcal{F}_i + \Phi_{\psi}(u) \approx \frac{\partial u_i}{\partial t},
\end{equation}
where $\Phi$ is parameterized by $\psi$. The outputs of State Kernels are then integrated by an ODE solver to solve for $u^{(t+1)}$, which will be used recursively for calculation of the subsequent time steps. The benefits of using an ODE solver are twofold: (a) it allows for adaptive time stepping, which in turn leads to better numerical stability in explicit schemes, and (b) it enables handling of unevenly spaced time series, which is very common in real observation data.

One of the benefits of State Kernels is to enable separate calculations for different quantities of interest, while the divergence (flux) can be calculated based on the same variable. This ensures that each quantity of interest follows its own conservation law. In short, FINN consists of Flux Kernels that handle the spatial correlation, and State Kernels that handle the temporal correlation of the data.

\section{Experiment}
For demonstration purposes, we choose an application with a subsurface contaminant transport problem.
We assess the performance of FINN not only using synthetic simulation data, but also real experimental data. The contaminant transport is characterized by the non-linear diffusion-sorption equation in a fluid-filled homogeneous porous medium:
\begin{equation}
    \label{eq:diff_sorp}
    R \frac{\partial c}{\partial t} = D_e \frac{\partial^2 c}{\partial x^2},
\end{equation}
where $c$ denotes the concentration of trichloroethylene (TCE) dissolved in the fluid, $D_e$ denotes the effective TCE diffusion coefficient, and $R$ denotes the retardation factor (representing sorption), which is a function of $c$. \autoref{eq:diff_sorp} is subject to a Dirichlet boundary condition at $x = 0$ and a Cauchy boundary condition at $x = L$ (i.e. the top and bottom ends of the field) and an initial condition $c(t=0) = 0$. Using the definition of retardation factors, we can also calculate the total TCE concentration $c_t$ (both in the fluid and adsorbed in the solid)
\begin{equation}
    \label{eq:c_tot}
    \frac{\partial c_t}{\partial t} = D_e \phi \frac{\partial^2 c}{\partial x^2},
\end{equation}
where $\phi$ is the porosity of the medium (i.e. the core samples). More detailed information on the experiment and its numerical simulation can be found elsewhere \citep{Nowak2016}.

\section{Results and Discussion}
As the first step, we train and test FINN using numerically generated synthetic datasets. Both train and test datasets are discretized into $26$ control volumes and $2\,000$ time steps. We train FINN using the whole spatial domain, with time steps $0 - 500$ (i.e. $t = 0 - 2 \, 500$ days) of the train dataset. Here, FINN receives only the initial condition, i.e. initial values of $c^{(0)}$ and $c_t^{(0)}$, along with the Dirichlet boundary condition value at the top boundary. The bottom boundary is subject to a Cauchy boundary condition, and therefore is solution dependent. FINN is then trained in a closed loop system, using predicted values of $c$ and $c_t$ at time step $t$ as input for the calculation at time step $t+1$.

For this synthetic data application, we set $\mathcal{N}$ and $\mathcal{D}$ to be learnable. Additionally, for the calculation of $c$, we set $\mathcal{D}$ to be a feedforward neural network that approximates $D_e/R$ in \autoref{eq:diff_sorp}, allowing us to extract information about the learned retardation factor as a function of the contaminant concentration $R(c)$. This neural network is constructed with 3 hidden layers, each containing 15 hidden nodes. Each hidden layer uses hyperbolic tangent as the activation function, and the output layer uses the sigmoid activation function, multiplied with a learnable scaling factor to ensure that the approximation of $D_e/R$ is strictly positive.

To test the generalization capability, we use the trained network to extrapolate until time step $2\,000$ ($t = 10 \, 000$ days). Additionally, we test the trained FINN prediction against a completely unseen test dataset obtained at different boundary conditions. The Dirichlet boundary condition values at the top boundary $c_s$ are $1.0$ kg/m\textsuperscript{3} and $0.7$ kg/m\textsuperscript{3} for the train and test dataset, respectively. We also compare the train and test Mean Squared Error (MSE) value with other known methods, such as TCN \citep{kalchbrenner2016neural}, ConvLSTM \citep{shi2015convolutional}, and DISTANA \citep{Karlbauer2019}.

\begin{table}[t!]
    \caption{Comparison of MSE values between different deep learning architectures.}
    \label{tab:benchmark_noise}
    \centering
    \footnotesize
        \begin{tabularx}{\linewidth}{lCCCc}
            \toprule
            Method & Training & Extrapolated training & Test unseen & Parameters \\
            \midrule
            TCN & $(7.9\pm5.4)\times10^{-6}$ & $(5.9\pm4.1)\times10^{-3}$  & $(3.0\pm1.2)\times10^{-2}$ & 1\,386\\
            ConvLSTM & $(5.5\pm1.6)\times10^{-6}$ & $(4.9\pm5.7)\times10^{-2}$ & $(6.6\pm7.9)\times10^{-2}$ & 1\,496\\
            DISTANA & $(1.9\pm1.1)\times10^{-6}$ & $(1.0\pm2.9)\times10^{-2}$ & $(1.6\pm4.0)\times10^{-2}$ & 1\,350\\
            FINN & $(4.7\pm4.9)\times10^{-5}$ & $(1.1\pm1.2)\times10^{-4}$ & $(4.1\pm4.0)\times10^{-5}$ & 528 \\
            \bottomrule
        \end{tabularx}
\end{table}

\begin{figure}[b!] 
     \centering
     \includegraphics[width=0.44\textwidth]{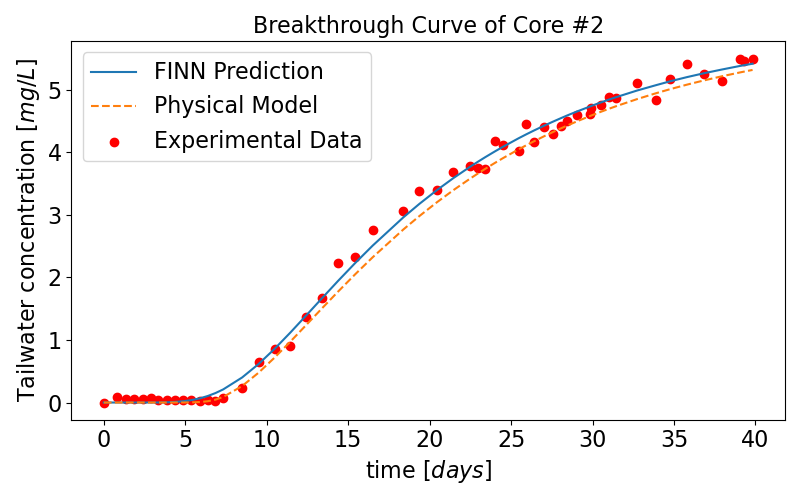}
     \hfill 
     \includegraphics[width=0.44\textwidth]{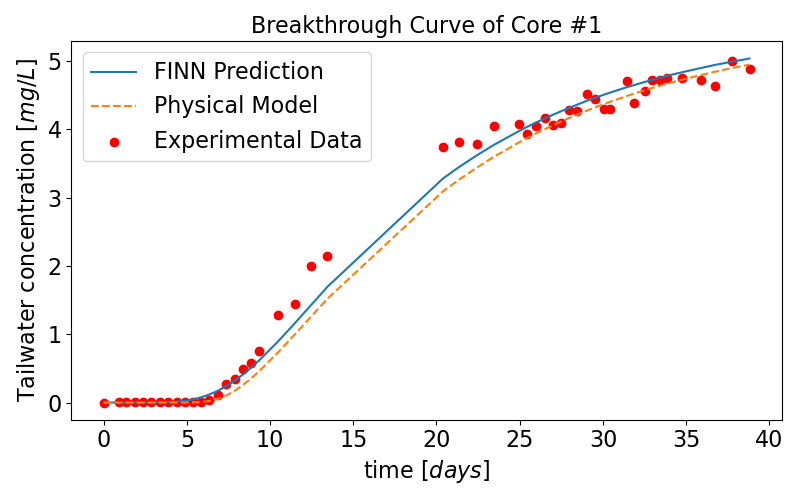}
     \\
     \includegraphics[width=0.44\textwidth]{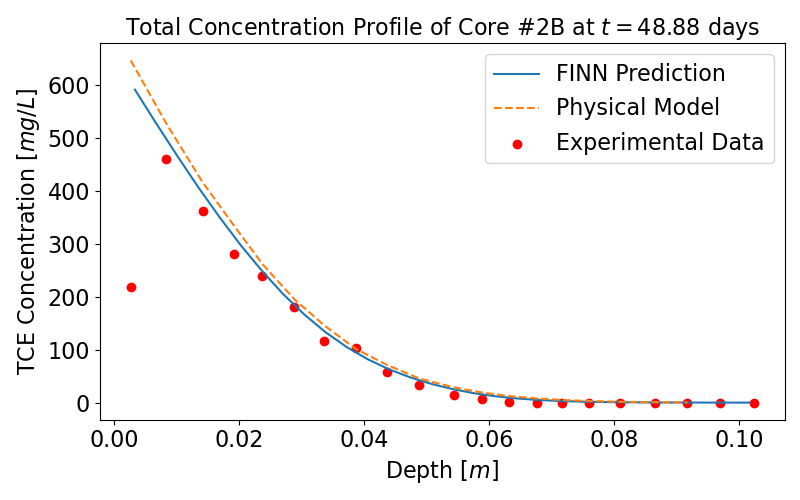}
     \hfill
     \includegraphics[width=0.44\textwidth]{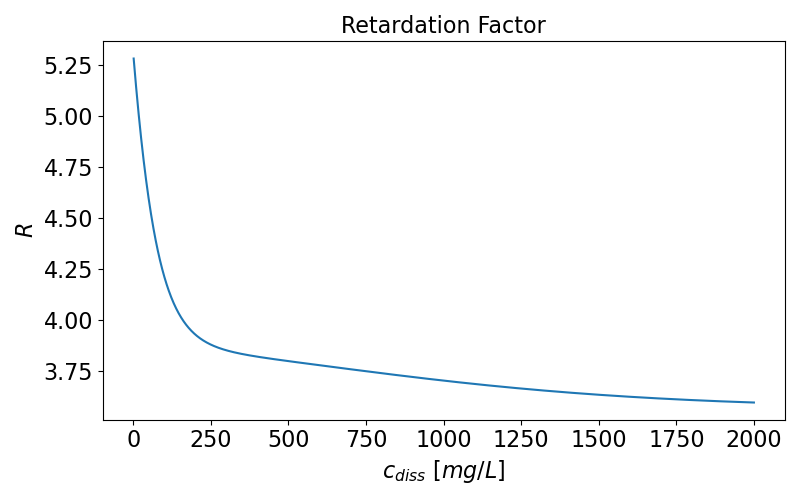}
    \caption{Breakthrough curve prediction of the FINN method (blue line) during training using data from core sample \#2 (top left), during testing using data from core sample \#1 (top right) and total concentration profile prediction using data from core sample \#2B (bottom left). The predictions are compared with the experimental data (red circles) and the results obtained using the physical model (orange dashed line). The extracted retardation factor as a function of $c$ is shown on the bottom right}
    \label{fig:experimental_data}
\end{figure}

The comparison in \autoref{tab:benchmark_noise} shows that FINN appropriately generalizes when tested against a different boundary condition. Even though all methods have comparable performance during training, the predictions of TCN, ConvLSTM, and DISTANA deteriorate when the knowledge gained from the training data needs to be extrapolated and to predict unseen test data. 
This appears to be caused by the fact that they lack proper boundary condition handling. More detailed information can be found in the appendices regarding the benchmark test dataset (\autoref{ap:data}), the training and test results (\autoref{ap:benchmark_plot}), as well as the model setup (\autoref{ap:model_details}).

As the second step, we apply FINN to real experimental data. The experimental data were collected from three different core samples, namely core samples \#1, \#2, and \#2B (see \autoref{ap:data_experiment}). In this setup, we train FINN using data that originate exclusively from core sample \#2. For this experimental data application, FINN is set up the same way as the synthetic data setup. For the dissolved concentration calculation, we also set $\mathcal{D}$ to be a feedforward neural network that takes $c$ as the input to learn the retardation factor function. However, we assume that we know the diffusion coefficient values for all core samples. The main setup difference lies in the available data used to train FINN. More specifically, we only use the breakthrough curve data of $c$ in the tailwater. This means that the data provides partial information and constrains the FINN prediction only at $x=L|_{0 \leq t \leq t_{end}}$.

The results show that the trained FINN has higher accuracy with ${\operatorname{MSE} = 4.84 \times 10^{-4}}$ compared to the calibrated (least squares) numerical PDE model with ${\operatorname{MSE} = 1.06 \times 10^{-3}}$. Further, we test and validate the trained model using different core samples (i.e. \#1 and \#2B), which originate from the same geographical area and therefore can be assumed to have similar soil parameters. In \autoref{fig:experimental_data}, we show that the predictions match the experimental data with reasonable accuracy. We also compare the predictions with the output of the numerical model, with the retardation factor also calibrated using the data from the same core sample \#2. The plots show that FINN's prediction accuracy is comparable to the numerical model, even beating it in some instances. For core sample \#1, the $\operatorname{MSE}$ of FINN prediction is ${1.37 \times 10^{-3}}$, while the numerical model underestimates the breakthrough curve with ${\operatorname{MSE} = 2.50 \times 10^{-3}}$.

Specifically for core sample \#2B, which is significantly longer than the other samples, to model the diffusion-sorption process, we can assume a zero-flux Neumann boundary condition at the bottom of the core. As a consequence, there are no breakthrough curve data available anymore. Instead, we compare the prediction against the total concentration profile data obtained from a destructive sampling (i.e. core slicing) at the end of the experiment. Here, FINN produces predictions with ${\operatorname{MSE} = 1.16 \times 10^{-3}}$, while the numerical model overestimates the total concentration profile with ${\operatorname{MSE} = 2.73 \times 10^{-3}}$. The results show that, even when applied to a different type of boundary condition, FINN's predictions remain accurate. Moreover, we can extract the retardation factor as a function of $c$ using FINN, which is plotted in \autoref{fig:experimental_data}, thus explaining a soil property.


\section{Conclusion and Future Work}
We have shown that including physical knowledge in the form of the Finite Volume structure produces excellent generalization capabilities and improves the explainability of the applied neural network structure. Our novel Finite Volume Neural Network (FINN) permits proper calculations for conservative fluxes and for different types of boundary condition. FINN outperforms other neural network methods for spatiotemporal modeling such as Temporal Convolutional Network, Convolutional LSTM, and DISTANA, especially when tested with different boundary conditions. Moreover, we show that FINN is suitable for experimental data processing, rendering it relevant as a data-driven modeling tool. 

In this work, we assumed spatial homogeneity for the soil in the simulation domain due to the small size of the actual experimental domain. For real field applications, where the scale is significantly larger, the homogeneity assumption might not hold. We are interested in enhancing FINN to handle spatially heterogeneous parameters. One way to achieve this is by defining a spatially correlated parameter to model a learnable diffusion coefficient in a spatially heterogeneous system akin to \cite{Karlbauer:2020b}. Additionally, we are also interested in quantifying uncertainties of our model by implementing a Bayesian Neural Network. A concrete application example in the contaminant transport modeling domain is to produce confidence intervals for both the concentration function and the learned retardation factor function. Overall, recent development in fusing numerical methods with deep learning to aid physical processes simulation shows promising results to keep continuing the trend in this direction. It is very exciting to see how far we can push the boundaries between numerical methods and deep learning and to see the benefit when combining both approaches.

\subsubsection*{Acknowledgments}
This work is funded by Deutsche Forschungsgemeinschaft (DFG, German Research Foundation) under Germany's Excellence Strategy - EXC 2075 – 390740016 as well as EXC 2064 – 390727645. We acknowledge the support by the Stuttgart Center for Simulation Science (SimTech). Moreover, we thank the International Max Planck Research School for Intelligent Systems (IMPRS-IS) for supporting Matthias Karlbauer. Codes and data that are used for this paper can be found in the repository \url{https://github.com/timothypraditia/finn}.

\bibliography{2021-ICLR-SimDL}
\bibliographystyle{iclr2021_conference}

\appendix
\section{Soil Parameters and Simulation Domains for the Benchmark Test}
\label{ap:data}
In this appendix, we present the soil parameters and the simulation domains used to generate the numerical benchmark dataset. \autoref{tab:data} summarizes the parameters used to generate the training and test dataset. The retardation factor function is generated with the Freundlich sorption isotherm, written mathematically as
\begin{equation}
    R = 1 + \frac{1-\phi}{\phi} \rho_s K_f n_f c^{n_f-1}.
\end{equation}

\begin{table}[h!]
    \caption{Soil parameters and simulation domains for training and testing dataset generation.}
    \label{tab:data}
    \centering
    \footnotesize
        \begin{tabularx}{0.8\linewidth}{cCccCc}
            \toprule
            \multicolumn{3}{c}{Soil parameters} & \multicolumn{3}{c}{Simulation domain} \\
            \midrule
            Parameter & Unit & Value & Parameter & Unit & Value \\
            \midrule
            $D_e$ & m\textsuperscript{2}/day & $5 \times 10^{-4}$ & $L$ & m & $1.0$\\
            $\phi$ & - & $0.29$ & $\Delta x$ & m & $0.04$\\
            $\rho_s$ & kg/m\textsuperscript{3} & $2880$ & $t_{end}$ & days & $10^4$\\
            $K_f$ & (m\textsuperscript{3}/kg)\textsuperscript{n\textsubscript{f}} & $3.53 \times 10^{-4}$ & $\Delta t$ & days & $5$\\
            $n_f$ & - & 0.874 & & &\\
            \bottomrule
        \end{tabularx}
\end{table}

            

Here, $D_e$ is the effective diffusion coefficient, $\phi$ is the porosity, $\rho_s$ is the bulk density, $K_f$ is the Freundlich K coefficient, $n_f$ is the Freundlich exponent, $L$ is the length of the sample, $\Delta x$ is the discrete control volume size, $t_{end}$ is the simulation time, and $\Delta t$ is the numerical time step.

For the training dataset, the upper boundary condition ($x=0$) is set to be a Dirichlet boundary condition, with the maximum solubility of TCE $c_s = 1.0$ kg/m\textsuperscript{3}. The testing dataset is generated with the same soil parameters and simulation domain, but with upper boundary condition $c_s = 0.7$ kg/m\textsuperscript{3}. The lower boundary condition ($x=L$) is set to be a Cauchy boundary condition according to $\frac{D_e}{Q} \frac{\partial c}{\partial x}|_{x=L}$, where $Q$ is the flow rate in the bottom reservoir. In the benchmark dataset, we assume that $Q=1.0$. Details on geometries, boundary conditions, and simulation can be found in \citep{Nowak2016}.

\section{Benchmark Test Results}
\label{ap:benchmark_plot}

In this appendix, we present the results and compare different methods for the benchmark test results. For each method, we train 10 models with different initialization. The MSE values of the predictions are then calculated compared with the training dataset at time steps $0 - 500$ (i.e. $t=0 - 2\,500$ days), the extrapolated training dataset at time steps $500 - 2 \, 000$ (i.e. $t = 2\,500 - 10\,000$ days), and the whole unseen test dataset (at all time steps $ 0 - 2\,000$). We train the models with noisy data. The noise is normally distributed with standard deviation $\sigma = 1 \times 10^{-5}$, i.e. $\mathcal{N}\sim(0.0, 1\times10^{-5}$). Detailed information of the test MSE for every individual model is shown in \autoref{ap:tab:mse_lists_seen_data} for seen data and in \autoref{ap:tab:mse_lists_unseen_data} for unseed data.

\begin{figure}[h!] 
     \centering
     \includegraphics[width=0.45\textwidth]{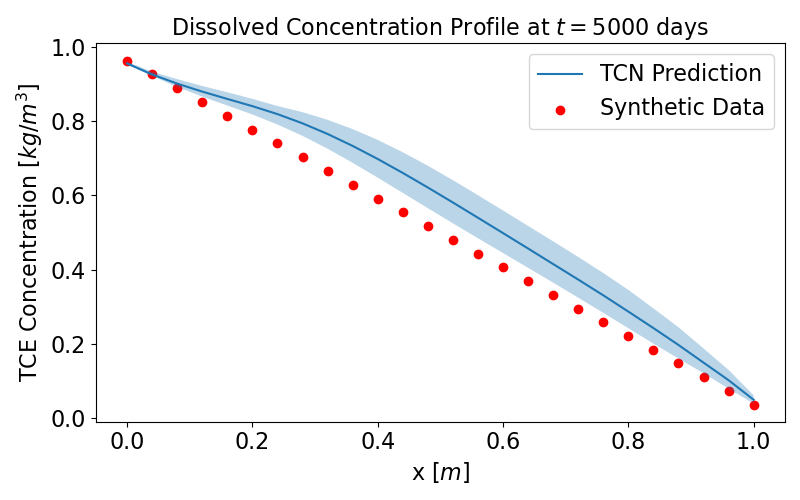}
     \hfill 
     \includegraphics[width=0.45\textwidth]{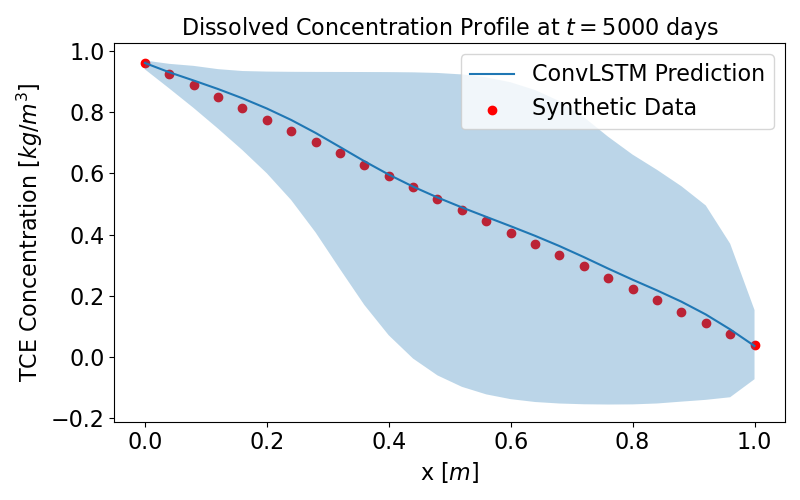}
     \\
     \includegraphics[width=0.45\textwidth]{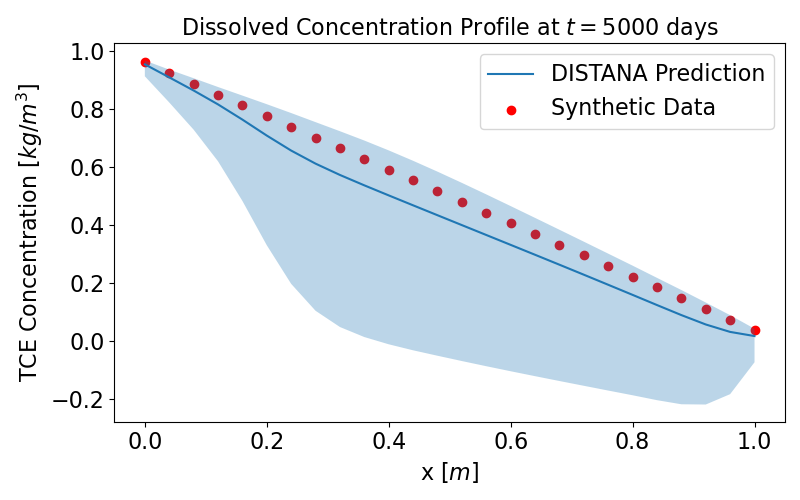}
    \hfill
     \includegraphics[width=0.45\textwidth]{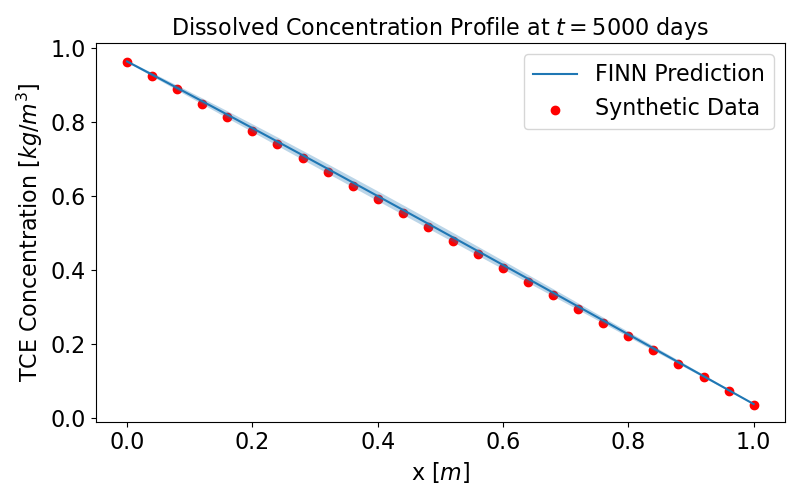}
    \caption{Dissolved concentration profile prediction mean (with confidence interval) at $t = 5\,000$ days compared with the extrapolated training dataset obtained using TCN (top left), ConvLSTM (top right), DISTANA (bottom left), and FINN (bottom right).}
        \label{fig:training_extrapolate_diss}
\end{figure}

\begin{figure}[h!] 
     \centering
     \includegraphics[width=0.45\textwidth]{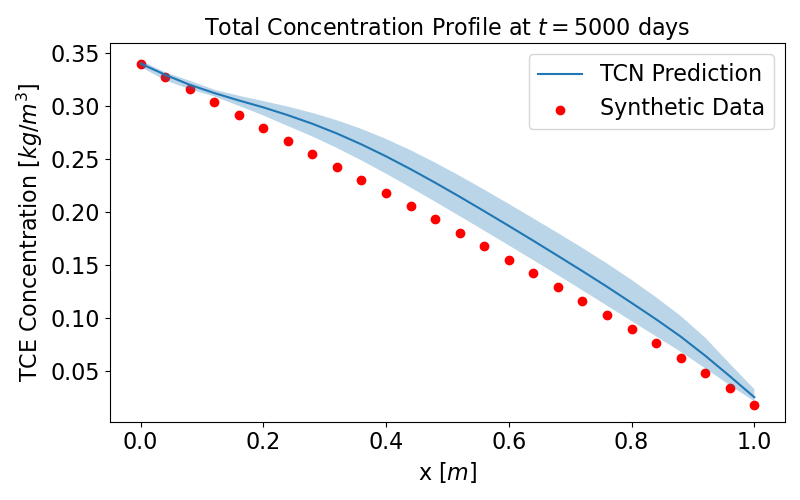}
     \hfill 
     \includegraphics[width=0.45\textwidth]{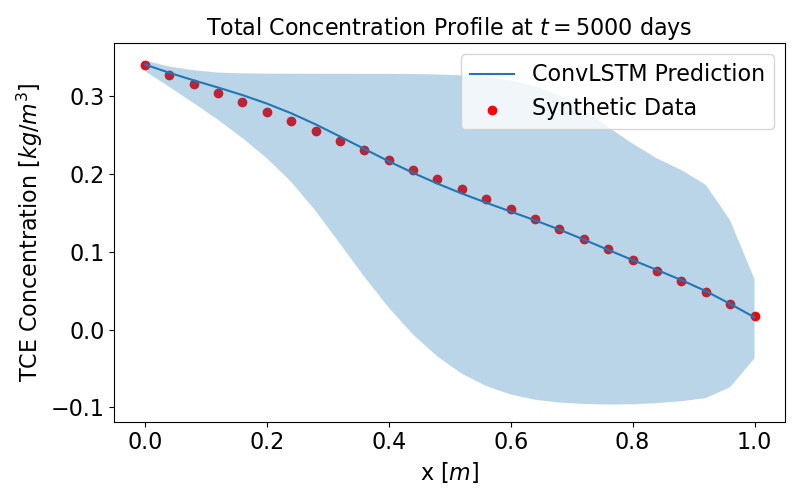}
     \\
     \includegraphics[width=0.45\textwidth]{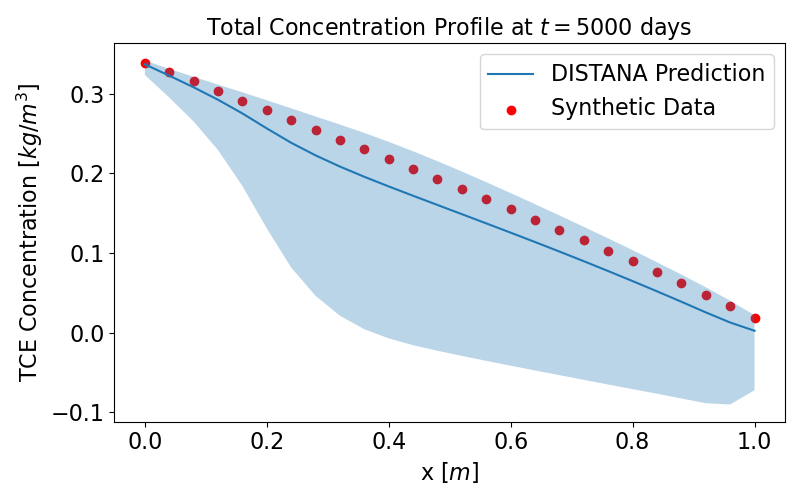}
     \hfill
     \includegraphics[width=0.45\textwidth]{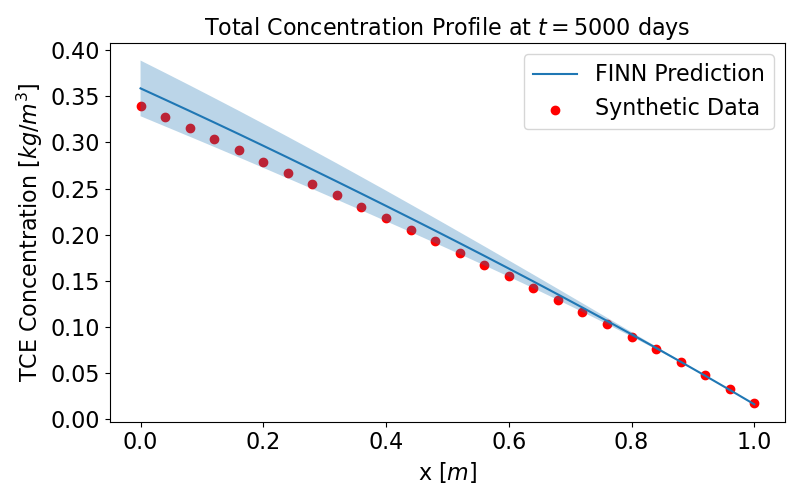}
    \caption{Total concentration profile prediction mean (with confidence interval) at $t = 5\,000$ days compared with the extrapolated training dataset obtained using TCN (top left), ConvLSTM (top right), DISTANA (bottom left), and FINN (bottom right).}
    \label{fig:training_extrapolate_tot}
\end{figure}

\begin{figure}[h!] 
     \centering
     \includegraphics[width=0.45\textwidth]{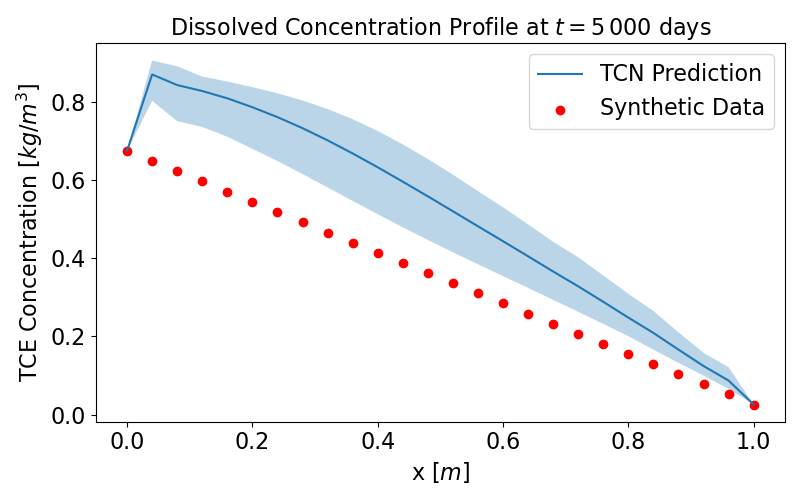}
     \hfill 
     \includegraphics[width=0.45\textwidth]{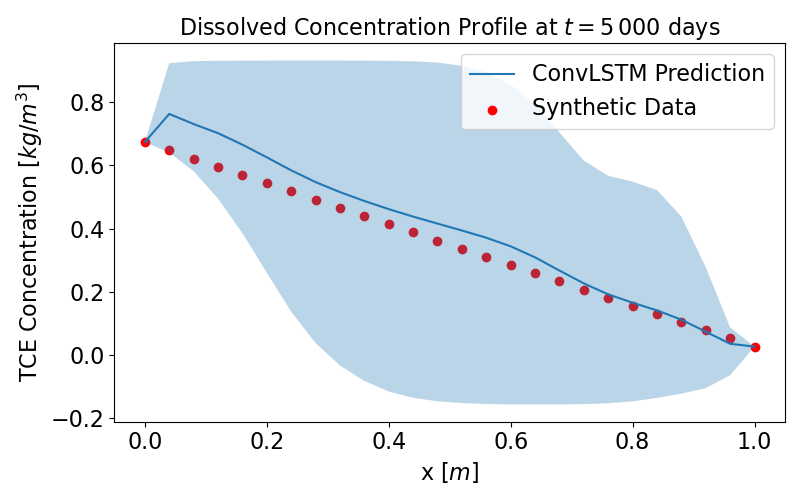}
     \\
     \includegraphics[width=0.45\textwidth]{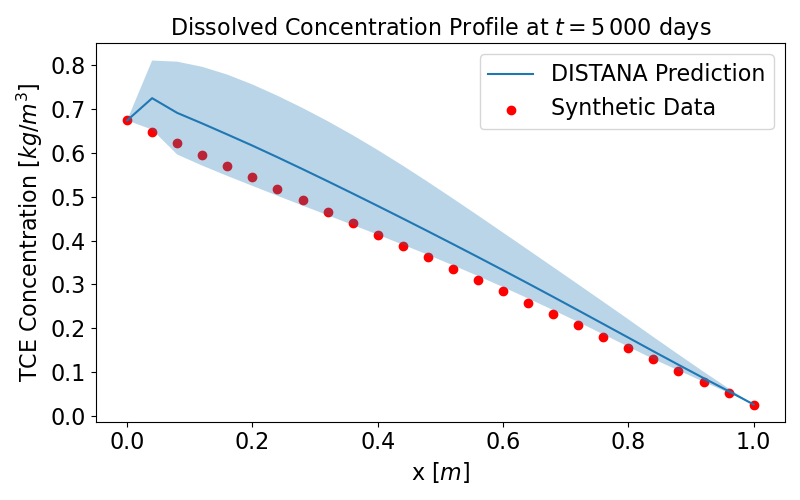}
     \hfill
     \includegraphics[width=0.45\textwidth]{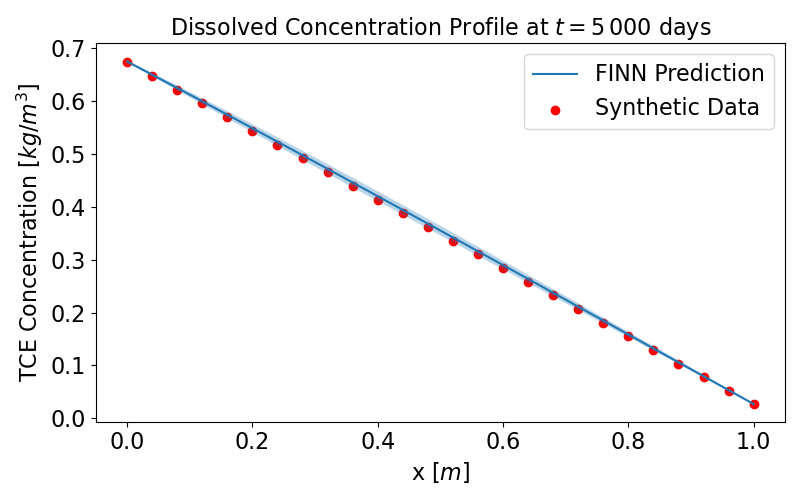}
    \caption{Dissolved concentration profile prediction mean (with confidence interval) at $t = 5\,000$ days compared with the test dataset obtained using TCN (top left), ConvLSTM (top right), DISTANA (bottom left), and FINN (bottom right).}
    \label{fig:test_diss}
\end{figure}

\begin{figure}[h!] 
     \centering
     \includegraphics[width=0.45\textwidth]{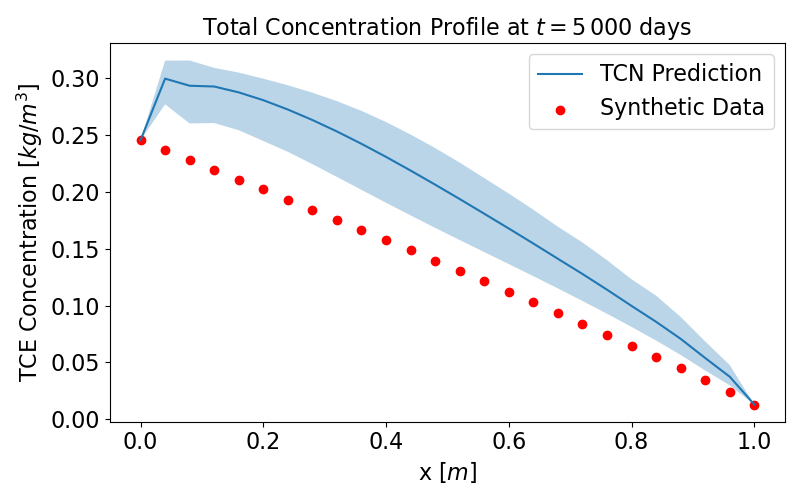}
      \hfill 
     \includegraphics[width=0.45\textwidth]{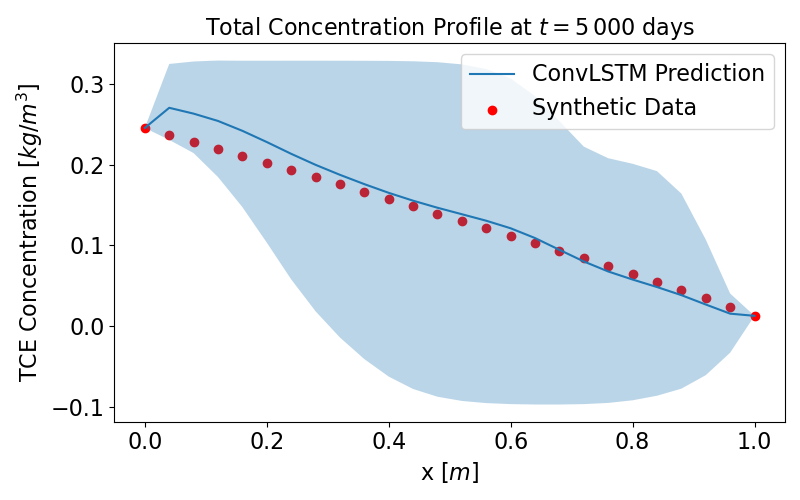}
     \\
     \includegraphics[width=0.45\textwidth]{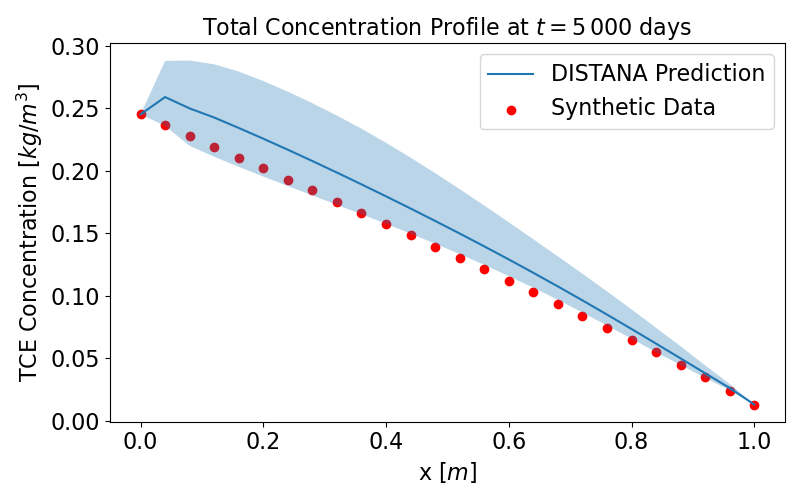}
     \hfill
     \includegraphics[width=0.45\textwidth]{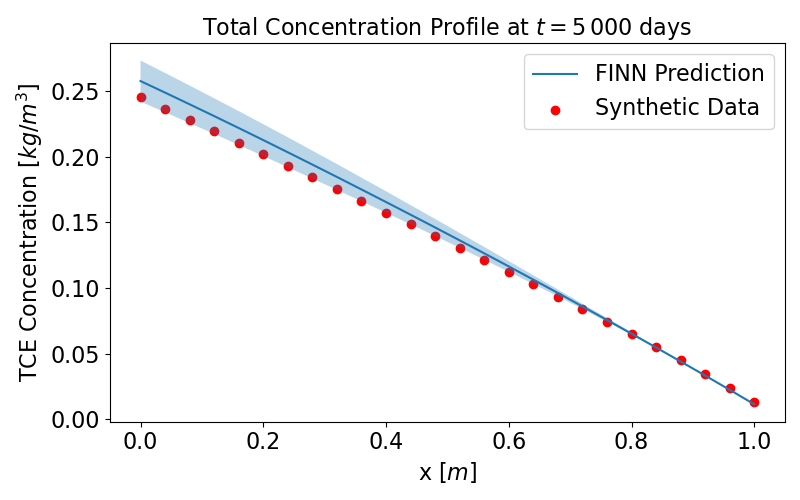}
     \caption{Total concentration profile prediction mean (with confidence interval) at $t = 5\,000$ days compared with the test dataset obtained using TCN (top left), ConvLSTM (top right), DISTANA (bottom left), and FINN (bottom right).}
    \label{fig:test_tot}
\end{figure}

The prediction mean and confidence interval are plotted in \autoref{fig:training_extrapolate_diss}, \autoref{fig:training_extrapolate_tot}, \autoref{fig:test_diss}, and \autoref{fig:test_tot}. Confidence intervals are obtained from repeated (ten times) training with random initialization. Even though the prediction mean of each method is not far from the synthetic data, clear instabilities and inconsistencies can be seen from the wide range of confidence intervals in the TCN, ConvLSTM, and DISTANA predictions. This instability is mainly caused by the improper handling of boundary conditions by these methods. FINN, on the other hand, produces very precise prediction along with high accuracy.


\begin{table}[h!]
    \caption{Test MSE on seen data (extrapolated training) from ten different training runs for each model}
    \label{ap:tab:mse_lists_seen_data}
    \centering
    
    \begin{tabularx}{\textwidth}{lCCCC}
        \toprule
        Run & TCN & ConvLSTM & DISTANA & FINN \\
        \midrule
        1  & $5.2\times10^{-3}$ & $2.1\times10^{-3}$ & $6.3\times10^{-5}$ & $2.7\times10^{-4}$ \\
        2  & $4.9\times10^{-3}$ & $3.4\times10^{-3}$ & $3.6\times10^{-4}$ & $3.0\times10^{-5}$ \\
        3  & $4.1\times10^{-3}$ & $8.3\times10^{-2}$ & $9.7\times10^{-2}$ & $2.7\times10^{-4}$ \\
        4  & $4.1\times10^{-3}$ & $1.5\times10^{-1}$ & $4.0\times10^{-4}$ & $9.0\times10^{-5}$ \\
        5  & $8.1\times10^{-4}$ & $4.4\times10^{-3}$ & $4.2\times10^{-5}$ & $8.6\times10^{-6}$ \\
        6  & $1.2\times10^{-2}$ & $8.3\times10^{-3}$ & $4.5\times10^{-4}$ & $4.1\times10^{-5}$ \\
        7  & $1.5\times10^{-2}$ & $2.4\times10^{-3}$ & $8.2\times10^{-5}$ & $3.2\times10^{-5}$ \\
        8  & $4.5\times10^{-3}$ & $5.0\times10^{-3}$ & $9.5\times10^{-4}$ & $2.8\times10^{-4}$ \\
        9  & $2.6\times10^{-3}$ & $1.0\times10^{-1}$ & $5.2\times10^{-5}$ & $2.4\times10^{-5}$ \\
        10 & $5.6\times10^{-3}$ & $1.3\times10^{-1}$ & $1.8\times10^{-4}$ & $3.5\times10^{-5}$ \\
        \bottomrule
    \end{tabularx}
\end{table}

\begin{table}[h!]
    \caption{Test MSE on unseen data coming from ten different training runs for each model}
    \label{ap:tab:mse_lists_unseen_data}
    \centering
    
    \begin{tabularx}{\textwidth}{lCCCC}
        \toprule
        Run & TCN & ConvLSTM & DISTANA & FINN \\
        \midrule
        1  & $3.8\times10^{-2}$ & $1.1\times10^{-2}$ & $1.5\times10^{-3}$ & $9.7\times10^{-5}$ \\
        2  & $3.3\times10^{-2}$ & $1.1\times10^{-3}$ & $8.9\times10^{-4}$ & $1.5\times10^{-5}$ \\
        3  & $3.0\times10^{-2}$ & $1.0\times10^{-1}$ & $1.4\times10^{-1}$ & $9.5\times10^{-5}$ \\
        4  & $2.7\times10^{-2}$ & $1.2\times10^{-1}$ & $8.6\times10^{-3}$ & $3.4\times10^{-5}$ \\
        5  & $2.5\times10^{-2}$ & $7.0\times10^{-3}$ & $7.0\times10^{-5}$ & $4.9\times10^{-6}$ \\
        6  & $5.1\times10^{-2}$ & $5.6\times10^{-4}$ & $3.6\times10^{-3}$ & $1.9\times10^{-5}$ \\
        7  & $2.9\times10^{-2}$ & $2.6\times10^{-2}$ & $3.0\times10^{-4}$ & $1.5\times10^{-5}$ \\
        8  & $3.9\times10^{-3}$ & $3.1\times10^{-4}$ & $8.6\times10^{-3}$ & $1.0\times10^{-4}$ \\
        9  & $2.3\times10^{-2}$ & $1.9\times10^{-1}$ & $3.4\times10^{-3}$ & $1.2\times10^{-5}$ \\
        10 & $4.3\times10^{-2}$ & $2.2\times10^{-1}$ & $3.7\times10^{-4}$ & $1.6\times10^{-5}$ \\
        \bottomrule
    \end{tabularx}
\end{table}

\section{Model Details of TCN, ConvLSTM and DISTANA}
\label{ap:model_details}
In this appendix, we provide additional information about the TCN, ConvLSTM and DISTANA models: bias neurons were used in all layers of all architectures and ADAM was used for optimization with a learning rate of $\eta=1\times10^{-3}$. As fair comparison, FINN results for the benchmark test case are obtained also using the ADAM optimizer with the same learning rate.
While TCN, ConvLSTM and DISTANA are always provided with the real data in the first ten timesteps (i.e. teacher forcing), FINN only receives an initial condition in the first timestep along with the information about the boundary in all timesteps.
For better comparison, we also provide boundary condition information for TCN, ConvLSTM, and DISTANA.
Note that in this experiment, TCN, ConvLSTM and DISTANA are provided with more information than FINN, which nevertheless outperforms the other models.

\paragraph{TCN} Two input channels are followed by two hidden layers with four and eight channels, respectively, which are processed by two output channels.
A convolution kernel size of $k=3$ was chosen and the standard dilation rate of TCN was applied ($d = l^2$, where $l$ is the index of the layer), leading to a time horizon of 28 time steps.
Code was taken and modified from \footnote{\url{https://github.com/locuslab/TCN}}.

\paragraph{ConvLSTM} Two input feature maps, followed by ten channels in one hidden layer and two output channels, were used. The convolution kernel size was set to $k=3$ and zero-padding was applied to preserve data dimensions.
PyTorch code was taken from \footnote{\url{https://github.com/ndrplz/ConvLSTM_pytorch}} and adapted to be applicable to spatially one-dimensional data.

\paragraph{DISTANA} Two input channels map to four preprocessing convolution channels, which feed forward into a ConvLSTM layer with eight channels which are processed by two postprocessing convolution channels.
The lateral information processing convolution layer was set to two channels.

\section{Soil Parameters and Simulation Domains for the Experiment}
\label{ap:data_experiment}
In this appendix, we present the soil parameters and the simulation domains of the core samples used in the experiment. \autoref{tab:data_experiment} summarizes the parameters of core sample \#1, \#2, and \#2B.

\begin{table}[h!]
    \caption{Soil and experimental parameters of core samples \#1, \#2, and \#2B.}
    \label{tab:data_experiment}
    \centering
    \footnotesize
        \begin{tabularx}{0.8\linewidth}{cCCCc}
            \toprule
            \multicolumn{5}{c}{Soil parameters} \\
            \midrule
            Parameter & Unit & Core \#1 & Core \#2 & Core \#2B \\
            \midrule
            $D_e$ & m\textsuperscript{2}/day & $2.00 \times 10^{-5}$ & $2.00 \times 10^{-5}$ & $2.78 \times 10^{-5}$\\
            $\phi$ & - & $0.288$ & $0.288$ & $0.288$\\
            $\rho_s$ & kg/m\textsuperscript{3} & $1957$ & $1957$ & $1957$\\
            
            \midrule
            \multicolumn{5}{c}{Simulation domain} \\
            \midrule
            Parameter & Unit & Core \#1 & Core \#2 & Core \#2B \\
            \midrule
            $L$ & m & $0.0254$ & $0.02604$ & $0.105$\\
            $r$ & m & $0.02375$ & $0.02375$ & N/A\\
            $t_{end}$ & days & $38.81$ & $39.82$ & $48.88$\\
            $Q$ & m\textsuperscript{3}/day & $1.01 \times 10^{-4}$ & $1.04 \times 10^{-4}$ & N/A\\
            $c_s$ & kg/m\textsuperscript{3} & $1.4$ & $1.6$ & $1.4$\\
            \bottomrule
        \end{tabularx}
\end{table}

For all experiments, the core samples are subjected to a constant TCE concentration at the top $c_s$, which amounts to a Dirichlet boundary condition. Notice that, for core sample \#2, we set $c_s$ to be slightly higher to compensate for the fact that there might be fractures at the top of core sample \#2, so that the TCE can break through the core sample faster.

For core samples \#1 and \#2, $Q$ is the flow rate of clean water at the bottom reservoir that determines the Cauchy boundary condition at the bottom of the core samples. For core sample \#2B, note that the sample length is significantly longer than the other samples. Therefore, for this particular sample, given $t_{end}$ to be approximately in the same order with the other samples, we assume the bottom boundary condition to be a no-flow Neumann boundary condition.

\end{document}